%% file: acl_latex.tex
\pdfoutput=1

\documentclass[11pt]{article}

\usepackage[preprint]{acl}

\usepackage{times}
\usepackage{latexsym}
\usepackage{graphicx}
\usepackage{tabularx}
\usepackage{fontawesome5}
\usepackage{amsmath}
\usepackage{amssymb, pifont}
\usepackage{multirow}
\usepackage{multicol}
\usepackage{array}
\usepackage{tcolorbox} 
\usepackage{fancyvrb}  
\usepackage{subcaption}
\usepackage{longtable}
\usepackage{caption}
\usepackage{colortbl}
\usepackage{booktabs} 
\usepackage{hyperref}
\usepackage[T1]{fontenc}

\newtcolorbox{mybox}[1][]{
    title=#1,
    fonttitle=\small,
    fontupper=\small,
    left=2mm,
    right=2mm,
    top=1mm,
    bottom=0mm,
}

\usepackage{subcaption}
\usepackage{graphicx}
\usepackage[utf8]{inputenc}
\usepackage{microtype}
\usepackage{float}
\usepackage{xspace}

\usepackage{inconsolata}
\usepackage{graphicx}
\newcommand\ours{\textsc{PersuasiveToM}\xspace}

\definecolor{yellow}{HTML}{F6BD60}
\definecolor{white}{HTML}{F7EDE2}
\definecolor{pink}{HTML}{F5CAC3}
\definecolor{tale}{HTML}{84A59D}
\definecolor{red}{HTML}{F28482}
\definecolor{green}{HTML}{8EC07C}
\definecolor{orange}{HTML}{FE8019}
\definecolor{grey}{HTML}{EBDBB2}
\definecolor{brain}{HTML}{FFABBE}
\definecolor{blue}{HTML}{076678}
\definecolor{narrative}{HTML}{458588}

\title{PersuasiveToM: A Benchmark for Evaluating Machine Theory of Mind \\ in Persuasive Dialogues}

\author{
    Fangxu Yu\textsuperscript{1} \quad Lai Jiang \textsuperscript{2} \quad Shenyi Huang \textsuperscript{3}  \quad Zhen Wu\textsuperscript{1}{\thanks{~~~Corresponding author.}} \quad {\bf Xinyu Dai}\textsuperscript{1} \\
    \textsuperscript{1}National Key Laboratory for Novel Software Technology, Nanjing University, China \\
    \textsuperscript{1}School of Artificial Intelligence, Nanjing University, China \\
    \textsuperscript{2}Department of Computer Science and Engineering, Shanghai Jiao Tong University \\
    \textsuperscript{3}University of California, San Diego, CA, USA \\
    {\tt yufx@smail.nju.edu.cn \quad jianglai0023-sjth@sjtu.edu.cn} \\
    {\tt {\tt shh058@ucsd.edu} \quad \{wuz, daixinyu\}@nju.edu.cn} 
}

\begin{document}
\maketitle
\begin{abstract}
The ability to understand and predict the mental states of oneself and others, known as the Theory of Mind (ToM), is crucial for effective social scenarios. Although recent studies have evaluated ToM in Large Language Models (LLMs), existing benchmarks focus on simplified settings (e.g., Sally-Anne-style tasks) and overlook the complexity of real-world social interactions. To mitigate this gap, we propose \ours, a benchmark designed to evaluate the ToM abilities of LLMs in persuasive dialogues. Our framework contains two core tasks: \textit{ToM Reasoning}, which tests tracking of evolving desires, beliefs, and intentions; and \textit{ToM Application}, which assesses the use of inferred mental states to predict and evaluate persuasion strategies.
Experiments across eight leading LLMs reveal that while models excel on multiple questions, they struggle with the tasks that need tracking the dynamics and shifts of mental states and understanding the mental states in the whole dialogue comprehensively. Our aim with \ours is to allow an effective evaluation of the ToM reasoning ability of LLMs with more focus on complex psychological activities. 
Our code is available at \href{https://github.com/Yu-Fangxu/PersuasiveToM}{https://github.com/Yu-Fangxu/PersuasiveToM}.
\end{abstract}

\section{Introduction}
Theory of Mind (ToM) involves the ability to reason about mental states both in oneself and in others~\cite{premack1978does}. This capacity strengthens many aspects of human cognition and social reasoning, enabling individuals to infer and simulate the mental states of others~\cite{gopnik1992child, baron1985does}. ToM is essential for various cognitive and social processes, including predicting actions~\cite{dennett1988precis}, planning based on others' beliefs and anticipated behaviors, and facilitating reasoning and decision making~\cite{pereira2016integrating, rusch2020theory}.

\begin{figure}[t]
\centering
\includegraphics[width=1.0\columnwidth]{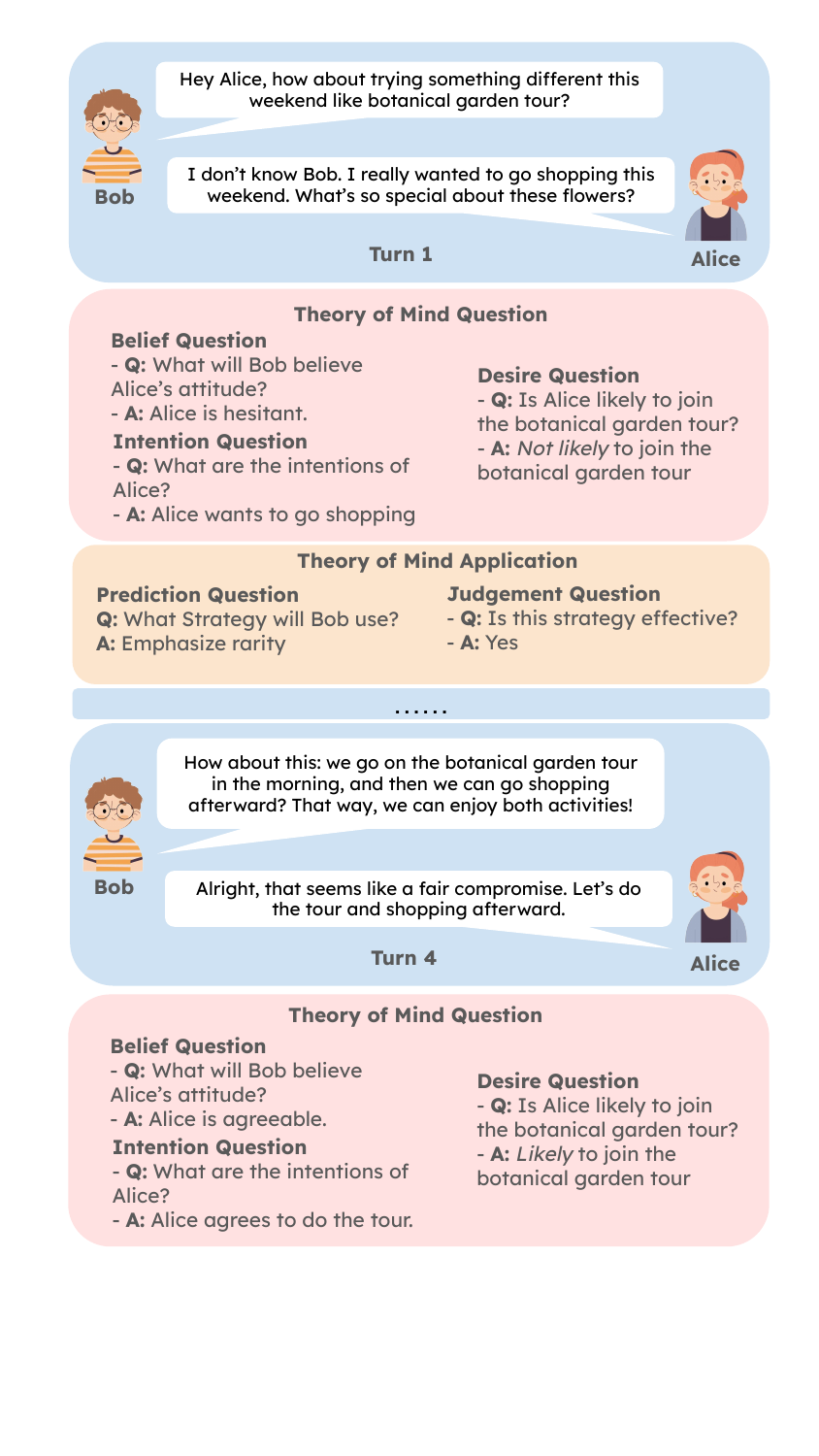}
\caption{An example in \ours. Bob is persuading Alice to join the botanical garden tour.}
\label{fig:example}
\vspace{-20pt}
\end{figure}

Recent advances in Large Language Models (LLMs) have demonstrated performance comparable to humans in problem-solving tasks. To assess whether LLMs exhibit high-level reasoning abilities regarding mental states, various studies have proposed benchmarks to evaluate their capacity to handle ToM tasks. A foundational concept in these benchmarks is the Sally-Anne test~\cite{baron1985does}, which has inspired the development of ToM evaluation frameworks~\cite{gu2024simpletom, he2023hi}. In this test, Anne secretly moves an object initially known to both Sally and Anne, leading Sally to hold a false belief about the object's location. The task requires participants to answer "Where will Sally look for the object?". Although this test assesses perception of the physical world, it fails to capture the complex dynamics of mental states in real-life social interactions and may not fully reflect ToM abilities in practical scenarios. To better simulate real-world social contexts, some benchmarks have been developed around communication scenarios~\cite{kim2023fantom, chan2024negotiationtom}. However, these benchmarks still focus primarily on inferring the scope of information regarding the physical world. This limits the ToM evaluation of understanding psychological states. Furthermore, current ToM benchmarks often overlook the critical step of applying ToM reasoning to predict actions, which is a key component of advanced social cognition. 

To address these limitations, we introduce \ours, a benchmark designed to evaluate LLMs' Theory of Mind capabilities specifically in realistic social interactions. Unlike previous benchmarks focused on inferring information about the physical world (e.g., object locations in Sally-Anne tests), \ours centers on understanding complex psychological states, such as a character's attitude towards an event. Inspired by the Belief-Desire-Intention (BDI) model \cite{bratman1987intention, georgeff1999belief}, \ours uses persuasive dialogue scenarios, characterized by asymmetric social status, to generate different psychological states for both parties. In addition, beyond assessing ToM reasoning, \ours evaluates the application of this understanding: assessing how well LLMs can predict actions (e.g., persuasive strategies) based on inferred mental states, and evaluating the effectiveness of persuasive strategies based on the persuadee's reactions.

Our evaluation results reveal several key findings: (1) LLMs score significantly lower than humans on questions requiring reasoning about dynamic changes (e.g., the persuadee's shifting desires) but perform competitively to humans on static aspects (e.g., the persuader's desires). (2) While Chain-of-Thought (CoT)~\cite{wei2022chain} prompting does not substantially improve performance on mental state reasoning, it enhances performance for most LLMs in predicting persuasion strategies. (3) LLMs exhibit distinct error patterns when reasoning about the persuader versus the persuadee, even when the question types are identical. (4) LLMs struggle to truly understand the dynamics of mental states of the whole dialogue, performing notably worse than humans in this regard.

\begin{table}[t]
    \centering
    \resizebox{1.0\columnwidth}{!}{ 
    \begin{tabular}{c| >{\centering\arraybackslash}p{1.0em} >{\centering\arraybackslash}p{1.0em} >{\centering\arraybackslash}p{1.0em} >{\centering\arraybackslash}p{1.0em} }
        \toprule 
        \multicolumn{1}{l}{{\color{yellow} \faUserFriends} : Social Interactions} & 
        \multicolumn{4}{l}{{\color{brain} \faBrain} : Psychological States } \\ 
        \multicolumn{1}{l}{{\color{green} \faBalanceScale} : Role Asymmetry}  &
        \multicolumn{4}{l}{{\color{blue} \faToolbox} : ToM Application}\\ \bottomrule
    \end{tabular}
    \vspace{-20pt}
    }
    \end{table}
    
    \begin{table}[t]
    \vspace{-10pt}
    \centering
    \resizebox{1.0\columnwidth}{!}{
    \begin{tabular}{c >{\centering\arraybackslash}p{1.6em} >{\centering\arraybackslash}p{1.6em} >{\centering\arraybackslash}p{1.6em} >{\centering\arraybackslash}p{1.6em} }
    \toprule 
    \noalign{\vspace{-8pt}}
         &
         {\color{yellow} \faUserFriends} &
         {\color{brain} \faBrain} &
         {\color{green} \faBalanceScale} &
         {\color{blue} \faToolbox}
         \\ \midrule 
        ToMi & \ding{56} & \ding{56} & \ding{56} & \ding{56} \\
        Hi-ToM & \ding{56} & \ding{56} & \ding{56} & \ding{56} \\
        ToMBench & \ding{56} & \ding{56} & \ding{56} & \ding{56} \\
        OpenToM & \ding{56} & \ding{52} & \ding{52} & \ding{56} \\
        Big-ToM & \ding{56} & \ding{56} & \ding{56} & \ding{52} \\
        FANToM & \ding{52} & \ding{56} & \ding{56} & \ding{56} \\
        NegotiationToM & \ding{52} & \ding{56} & \ding{56} & \ding{52} \\
        \midrule 
        \ours & \ding{52} & \ding{52} & \ding{52} & \ding{52} \\
        \bottomrule 
    \end{tabular}
    }
    \vspace{-0.75em}
    \caption{Comparison of \ours with existing ToM datasets. In the header, \textit{Psychological ToM} refers to testing ToM abilities in characters' mental states of the psychological activities. }
    \vspace{-1em}
    \label{tab:benchmark_comparison}
\end{table}
\begin{table*}[t]
\centering
\resizebox{0.95\textwidth}{!}{%
\begin{tabular}{c|l}
\toprule
\textbf{Type} & \textbf{\ours Questions} \\
\midrule
Desire Question & Is <Persuader/Persuadee> likely to <Target of Persuasion> ?\\
\midrule
Belief Question & What will <Persuader/Persuadee> believe <Persuadee/Persuader>'s attitude towards <Target of Persuasion> ?\\
\midrule
Intention Question & What are the intentions of <Persuader/Persuadee> expressed in <Utterance> given the dialogue history? \\
\midrule
Prediction Question & What strategy will the persuader use next? \\
\midrule
judgement Question & <Persuader> will adopt <Strategy> to persuade <Persuadee> to <Target of Persuasion>. Is this strategy (not) effective? \\
\bottomrule
\end{tabular}
}
\vspace{-5pt}
\end{table*}

\begin{table*}[t]
\centering
\resizebox{0.95\textwidth}{!}{%
\begin{tabular}{l|l|l|l|l|l}
\toprule
 & \multicolumn{3}{c|}{\textbf{ToM Reasoning}} & \multicolumn{2}{c}{\textbf{ToM Application}} \\\midrule
 \textbf{Dialogue} & \textbf{Desire} & \textbf{Belief} & \textbf{Intention} &\textbf{Prediction} & \textbf{Judgement} \\\midrule
 \multicolumn{6}{c}{\textbf{1st Turn Dialogue}} \\\midrule
\textbf{Bob says:} Hey Alice, how about trying something   & Persuade Alice to  & Alice is  & Intent to make the other  & Emphasize rarity & Strategy "Emphasize  \\
different this weekend? The botanical garden tour  & join the botanical & hesitant & person feel the experience or & & rarity" is Effective\\
is a unique experience, and you'll get to take & garden tour & &  objects are unique or scarce. & \\
 stunning pictures of the exotic flowers on display! & & & &  \\\midrule
\textbf{Alice says:} I don't know, Bob. I really wanted to  & Not likely to join the & Bob is & Alice wants to go shopping. & & \\
go shopping this weekend. What's so special  & botanical garden tour & enthusiastic & & \\
about these flowers? & & & &  \\\midrule
 \multicolumn{6}{c}{\textbf{2nd Turn Dialogue}} \\\midrule
\textbf{Bob says:} These flowers are incredibly rare, and & Persuade Alice to  &  Alice is & Intent to make the other  & Mention garden's & Strategy "Mention  \\
it's not often that you get to see such a diverse and   & join the botanical & curious & person feel the experience or & history & garden's history" is \\
exotic collection up close. Some of these plants are  & garden tour & & objects are unique or scarce. & & effective\\
not found anywhere else in the world! & & & &  \\\midrule

\textbf{Alice says:} Really? That sounds interesting, but & Neutral to join the  & Bob is & Alice is curious but not  & & \\
I'm not sure if it's worth giving up shopping for. & botanical garden tour & excited & yet convinced. & & \\\midrule
 \multicolumn{6}{c}{\textbf{3rd Turn Dialogue}} \\\midrule
\textbf{Bob says:} The botanical garden has a rich history, & Persuade Alice to  & Alice is & Intent to demonstrating the  & Suggest shopping & Strategy "Suggest \\
and the expert guides can teach us so much about & join the botanical & considering & expertise of the domain & afterward & shopping afterward"  \\
the plants and their unique stories. Plus, it's a great  &  garden tour & &  and showing authority. & & is effective\\
opportunity to learn something new while enjoying & & & & \\
nature's beauty. & & & & \\\midrule
\textbf{Alice says:} Hmm, that does sound intriguing, but I  & Not likely to join the & Bob is & Alice is considering the idea  & & \\
still want to go shopping.&  botanical garden tour & informative & but still wants to shop. &  \\\midrule
  \multicolumn{6}{c}{\textbf{...}} \\
\bottomrule
\end{tabular}
}
\caption{An example dialogue from the \ours benchmark, illustrates the tracking of mental states (desire, belief, intention) and the application of ToM reasoning in predicting and evaluating persuasion strategies across multiple turns. The upper part contains questions in the \ours benchmark.}
\vspace{-10pt}
\label{tab:main_arch}
\end{table*}

\section{Related Works}
\paragraph{Theory of Mind Benchmarks.} Existing ToM evaluation benchmarks for LLMs are mostly text story-based QA forms \cite{gandhi2024understanding, le2019revisiting, kim2023fantom, he2023hi, gu2024simpletom}, with some multi-modal extensions \cite{jin2024mmtom, shi2024muma}, which adapt the Sally-Anne test~\cite{baron1985does}. These benchmarks ask models to determine the true beliefs or knowledge of individuals based on a given premise~\cite{sclar2023minding, ullman2023large, shapira2023clever, ma2023towards}. Story-based benchmarks focus on reasoning about mental states of the physical world without on psychological states, and applying ToM for decision-making in real-world social interactions.
To address this, \cite{hou2024entering} evaluates ToM in situated environments, and \cite{chan2024negotiationtom} focuses on negotiation scenarios. However, the latter is limited to bargaining specific resources, while ToMATO~\cite{shinoda2025tomato} uses persuasive dialogues but focuses solely on donation requests, limiting its diversity. For decision-making, BigToM~\cite{gandhi2024understanding} assesses action prediction in narratives, but without grounding in interactive social interactions.
Persuasion involves complex psychological dynamics and asymmetric relationships, providing a rich testbed for assessing ToM in social interactions. Built upon persuasive dialogues, \ours is designed to evaluate an LLM's ability to reason about the mental states of individuals and apply this understanding to predict and assess persuasion strategies, bridging ToM reasoning with decision-making in social interactions. Table~\ref{tab:benchmark_comparison} compares \ours with existing ToM benchmarks.

\paragraph{Persuasive Dialogue.} Persuasive dialogues aim to influence the beliefs, attitudes, or behaviors of individuals through communication strategies \cite{shi2020effects}. Recent works have tried to develop datasets or facilitate LLMs with persuasion techniques to achieve specific goals. Previous datasets are constructed by crowd-sourcing \cite{wang2019persuasion} or synthesizing with LLMs~\cite{zhou2023sotopia, jin2024persuading}. Many of the previous works build an effective persuasive dialogue system from emotional influence \cite{samad2022empathetic}, social facts \cite{chen2022seamlessly}, and strategies \cite{tian2020understanding, jin2023joint}.
Previous persuasion techniques have been widely adopted to jailbreak LLMs \cite{zeng2024johnny}, mislead LLMs \cite{xu2023earth}, as well as for information retrieval \cite{furumai2024zero}. A similar work \cite{sakurai2024evaluating} evaluates the intention detection abilities of LLMs in persuasive dialogues; however, \ours introduces a more comprehensive benchmark to assess the ToM abilities of LLMs in such contexts. In addition, we discuss the practical connection with dialogue generation and personalization in Appendix~\ref{discussion: dialogue}.

\section{\ours Benchmark}

\subsection{Overview}
We construct \ours benchmark to evaluate the Theory of Mind (ToM) abilities of LLMs in dynamic, multi-turn persuasive dialogues with asymmetric social status, which give rise to distinct mental states. \ours focuses on two core dimensions: \textit{ToM Reasoning} (§\ref{sec: tom reasoning}), assessing whether models can track evolving desires, beliefs, and intentions of both persuader and persuadee; and \textit{ToM Application} (§\ref{sec: tom application}), evaluating whether LLMs can use these inferred states to predict or assess persuasion strategies. Kye design considerations include: (1) coverage of diverse domains (e.g., \textit{life, education, technology, etc.}) to ensure a comprehensive evaluation in the social context. (2) evolving mental states across multi-turn interactions to assess whether LLMs can track the shift in the dialogue. (3) asymmetric roles have different mental states (e.g., stable goals for persuaders vs. shifting states for persuadees under active persuasion).

Table~\ref{tab:main_arch} shows a \ours example, illustrating how mental states—desires, beliefs, and intentions—are tracked and inferred across turns, and how they inform the prediction and evaluation of persuasion strategies. This highlights the dynamic nature of real-world persuasion and the reasoning challenges it poses for LLMs. 
\subsection{Data Source}
\ours is annotated on the multi-turn persuasive dialogue dataset DailyPersuasion~\cite{jin2024persuading}. Each instance in DailyPersuasion is an $N$-round alternating dialogue $D = [(u_1^{a}, u_1^{b}, s_1^{a}), (u_2^{a}, u_2^{b}, s_2^{a}), ..., (u_N^{a}, u_N^{b}, s_N^{a})]$ between the persuader $a$ and the perusadee $b$, and accompanied with a persuasion strategy $s_i^{a}$. Persuadee $b$ has a different desire from $a$ initially, after multi-turn persuasion, persuadee $b$ changes the mind to agree or consider the proposal of $a$. 
\subsection{ToM Reasoning}
\label{sec: tom reasoning}
In \ours, we break down ToM reasoning into three core reasoning tasks: \textit{Desire Reasoning}, \textit{Belief Reasoning}, and \textit{Intention Reasoning} for evaluation, which matches Belief-Desire-Intention (BDI) modeling~\cite{bratman1987intention}. Questions are listed in Table~\ref{tab:main_arch}.
\vspace{-8pt}
\paragraph{Desire Reasoning.} Desire represents a motivational state that drives behavior but does not necessarily imply a firm commitment~\cite{malle2001distinction, kavanagh2005imaginary}. Desires are seen as either fulfilled or unfulfilled which is different form beliefs that are evaluated in terms of truth or falsity. In \ours, we evaluate LLMs’ ability to comprehend and track the evolution of desires in both persuaders and persuadees. 
For the persuader, the desire is typically static, representing their goal (e.g., persuade Alice to join the botanical garden tour). For the persuadee, however, desires are dynamic and shift in response to the persuader’s tactics (e.g., Alice’s initial desire to shop transforms into a willingness to compromise). To assess this, we design \textit{Desire Questions} that probe two key aspects: (1) Can LLMs consistently identify the persuader's static desire throughout the dialogue? (2) Can LLMs track the dynamics of the persuadee's desire shifting from refusal or disinterest to being persuaded? For evaluation, we annotate the persuader’s desire questions as the persuasive goal in DailyPersuasion and use LLMs to annotate whether the persuadee is ultimately persuaded. See Appendix~\ref{sec: data anno} for details.

\begin{table}[t]
\centering
\resizebox{0.9\columnwidth}{!}{%
\begin{tabular}{l|l}
\toprule
\parbox{2cm}{\textbf{Principles}} & \textbf{Intentions} \\ \midrule
\parbox{2cm}{Reciprocity} & \parbox{5cm}{Make the other person feel accepted through concessions, promises, or benefits.} \\ \midrule
\parbox{2cm}{Scarcity} & \parbox{5cm}{Make the other person feel the experience or objects are unique or scarce.} \\ \midrule
\parbox{2cm}{Consensus} & \parbox{5cm}{Refer to what other people are doing, or what they have already purchased or done.} \\ \midrule
\parbox{2cm}{Authority} & \parbox{5cm}{Demonstrating expertise of the domain and showing authority.} \\ \midrule
\parbox{2cm}{\raggedright Commitment \& Consistency}  & \parbox{5cm}{Encourage the other person to commit to take the first step and be consistent.} \\ \midrule
\parbox{2cm}{Liking} & \parbox{5cm}{Praising other people or finding common characteristics to improve the other person's liking.} \\ \bottomrule
\end{tabular}
}
\caption{Intention mapping from the persuasive principles. Refer to Appendix~\ref{sec: persuasion} for definitions of persuasive principles.}
\label{intention}
\vspace{-10pt}
\end{table}
\vspace{-8pt}
\paragraph{Belief Reasoning.} Belief is a cognitive state where an individual holds a particular perspective, attitude, or viewpoint regarding a given proposition or idea. In \ours, beliefs refer to understanding and reasoning the attitudes of the opponent toward the goal, which is explicitly or implicitly expressed in the dialogue. For example, in Turn 1, Bob believes Alice is hesitant about the tour, while Alice believes Bob is enthusiastic. By Turn 3, Bob’s belief shifts to thinking Alice is considering the idea, while Alice becomes more informed about the garden’s history. \textit{Belief Questions} ask LLMs to infer what will <persuader/persuadee> believe <persuadee/persuader>'s attitude towards the persuasion goal. These questions require models to understand cues in utterances and update beliefs dynamically as the dialogue progresses. We annotate the attitudes as the tone of each utterance of both persuaders and persuadees in DailyPersuasion. 
\vspace{-8pt}
\paragraph{Intention Reasoning.} Intentions represent deliberate commitments to pursue specific goals based on desires and beliefs, often linked to tangible actions aimed at achieving those objectives. Intentions have been extensively studied in psychology tests such as action prediction~\cite{phillips2002infants} and behavioral re-enactment~\cite{meltzoff1995understanding}. Inspired by persuasion principles~\cite{cialdini2004social, cialdini2007influence}, we develop a mapping from persuasion principles to intentions, as shown in Table~\ref{intention}.
In persuasive dialogue, persuasive strategies have a strong association with intentions~\cite{wang2019persuasion}. In \ours, we collect the persuasive strategies from the DailyPersuasion dataset and their corresponding utterances for prompting the LLMs to choose the most appropriate intentions from table~\ref{intention}. The details of the extraction are recorded in Appendix~\ref{sec: data anno}.
For the persuader, we ask LLMs to choose the most appropriate intention from the six designed intention choices. For the persuadees, intentions are summarized and extracted by LLMs from their utterances.

\subsection{ToM Application}
\label{sec: tom application}
While ToM reasoning plays a crucial role, it is equally important to analyze how LLMs utilize the understanding of mental states to proactively influence others' thoughts and decisions. To this end, we propose to assess LLMs' ability to leverage the understanding of mental states in a dialogue for identifying the most effective persuasive strategies and evaluating the effectiveness of persuasive strategies based on the persuadee's response. These tasks test whether LLMs can leverage inferred mental states to guide strategic decision-making, bridging the gap between reasoning and action.

\begin{figure}[t]
\centering
\includegraphics[width=0.80\columnwidth]{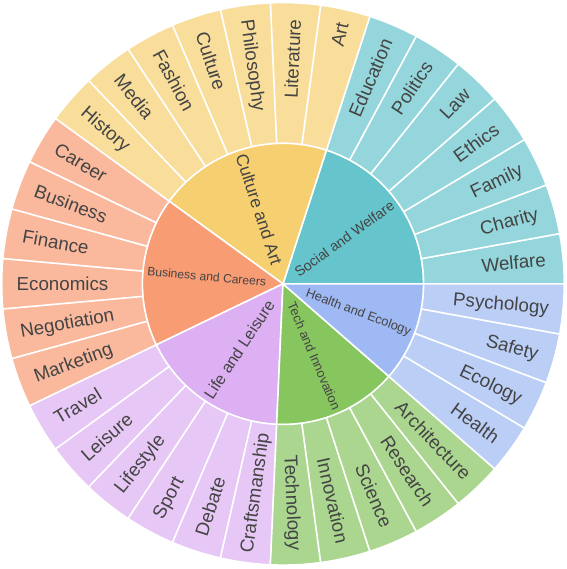}
\caption{Domains of \ours. Under 6 primary topics and 35 domains in total.}
\vspace{-10pt}
\label{fig: domain}
\end{figure}

\paragraph{Persuasion Strategy Prediction.} This question involves asking which persuasion strategy the persuader is likely to employ next from a set of possible strategies. To answer these questions correctly, LLMs need to reason over the dialogue to infer the mental states of characters and predict what the likely next prediction strategy is to further influence the persuadee's beliefs, desires, and intentions, ultimately achieving the desired persuasion outcome.
\vspace{-10pt}
\paragraph{Judgement Question.}
The judgment question specifies that the correct strategy was taken, and asks LLMs if the selected strategy is effective for persuasion. Answering such questions requires reasoning about the beliefs and intentions of the persuadee. Only by accurately inferring the persuadee's mental state can one properly determine whether the persuasion strategies should be employed to convince the persuadee.
\subsection{Statistics}
In Table~\ref{tab: statis}, we present the data statistics of \ours. As shown in Figure~\ref{fig: domain}, \ours includes diverse domains. These real-life domains are crucial for comprehensively evaluating LLMs in social interactions. We sample 15 dialogues from each domain to form the dataset in \ours. We create multi-choice questions by either prompting GPT-4o~\cite{hassany2025generating} to generate semantically different choices or randomly selecting three distractors from the predefined pool. Refer to Appendix~\ref{sec: data anno} for more details. 

\begin{table}[t]
\centering
\resizebox{0.75\columnwidth}{!}{%
\begin{tabular}{lc}
\toprule
\# Domains & 35 \\
\# Dialog instances & 525 \\
\# Avg. Turns Per Dialog & 4.9 \\
\# Avg. Words Per Turn & 61.3 \\ \midrule
\multicolumn{2}{c}{\textit{Questions}} \\\midrule
\# Desire (er/ee) & 2568/2459 \\
\# Belief (er/ee) & 2580/2580 \\
\# Intention (er/ee) & 2568/2041 \\
\# Strategy prediction & 2041 \\
\# Strategy judgement & 2041 \\
\bottomrule
\end{tabular}
}
\caption{Statistics of \ours dataset.}
\label{tab: statis}
\vspace{-15pt}
\end{table}

\begin{table*}[t]
\centering
\resizebox{1.0\textwidth}{!}{%
\begin{tabular}{l|ccc|ccc|cc}
\toprule
~& \multicolumn{6}{c|}{\textbf{ToM Reasoning}} & \multicolumn{2}{c}{\textbf{ToM Application}}\\ \midrule
~& \multicolumn{3}{c|}{\textbf{Persuader}} & \multicolumn{3}{c|}{\textbf{Persuadee}} & \\\midrule
Model & Desire & Belief & Intention & Desire & Belief & Intention & Strategy Pred & Judgement \\
\midrule
Random Guess & 50.00 & 25.00 & 16.67 & 33.33 & 25.00 & 25.00 & 25.00 & 50.00\\
Human & 100.00 & 92.31 & 78.13 & 84.62 & 87.85 & 94.42 & 86.80 & 97.95   \\
\midrule
LLaMa-3.1-8B-Instruct       &   59.78$_{\pm 1.41}$ &  65.84$_{\pm 0.63}$ &  41.45$_{\pm 0.85}$ &   68.09$_{\pm 2.57}$ &  71.54$_{\pm 0.40}$ &  83.67$_{\pm 4.38}$ &  61.69$_{\pm 0.76}$ &   93.58$_{\pm 3.95}$ \\
Qwen2.5-7B-Instruct         &   91.08$_{\pm 7.42}$ &  83.09$_{\pm 1.01}$ &  \underline{46.09}$_{\pm 0.63}$ &   64.76$_{\pm 0.55}$ &  79.00$_{\pm 0.08}$ &  82.77$_{\pm 4.35}$ &  63.23$_{\pm 6.06}$ &   \underline{95.63}$_{\pm 0.28}$ \\
Gemma-2-9b-it               &   \textbf{96.06}$_{\pm 3.02}$ &  \underline{83.36}$_{\pm 1.20}$ &  45.42$_{\pm 0.78}$ &   62.87$_{\pm 2.12}$ &  64.62$_{\pm 0.78}$ &  80.24$_{\pm 2.93}$ &  64.75$_{\pm 0.39}$ &   63.23$_{\pm 8.92}$ \\
GLM4-9B-Chat                &   87.84$_{\pm 2.28}$ &  69.00$_{\pm 6.55}$ &  40.82$_{\pm 0.59}$ &   63.18$_{\pm 2.90}$ &  67.25$_{\pm 0.61}$ &  83.54$_{\pm 2.56}$ &  61.10$_{\pm 0.27}$ &   93.38$_{\pm 2.70}$ \\
Mixtral-8x7B-Instruct       &   93.10$_{\pm 0.59}$ &  71.03$_{\pm 4.90}$ &  41.68$_{\pm 1.80}$ &   68.27$_{\pm 2.08}$ &  70.09$_{\pm 3.50}$ &  83.92$_{\pm 1.67}$ &  63.06$_{\pm 2.70}$ &   95.36$_{\pm 2.05}$ \\
InternLM-2.5-7B-Chat        &   83.73$_{\pm 0.09}$ &  70.46$_{\pm 1.14}$ &  39.77$_{\pm 0.13}$ &   \textbf{71.94}$_{\pm 0.69}$ &  69.50$_{\pm 0.54}$ &  85.29$_{\pm 2.86}$ &  64.01$_{\pm 1.50}$ &  74.29$_{\pm 0.61}$ \\
GPT-4o-mini                 &   94.70$_{\pm 0.22}$ &  70.66$_{\pm 0.96}$ &  45.19$_{\pm 0.66}$ &   70.78$_{\pm 0.09}$ &  \underline{78.25}$_{\pm 2.52}$ &  \underline{86.62}$_{\pm 1.33}$ &  \underline{66.13}$_{\pm 0.90}$ &   91.52$_{\pm 1.60}$ \\
GPT-4o                      &   \underline{95.56}$_{\pm 4.51}$ &  \textbf{89.47}$_{\pm 0.02}$ &  \textbf{46.28}$_{\pm 0.37}$ &  67.66$_{\pm 2.23}$ &  \textbf{81.73}$_{\pm 2.41}$ &  \textbf{87.81}$_{\pm 1.38}$ &  \textbf{73.55}$_{\pm 0.78}$ &   \textbf{96.38}$_{\pm 1.66}$ \\\midrule
LLaMa-3.1-8B-Instruct + CoT &   59.83$_{\pm 0.11}$ &  68.91$_{\pm 1.27}$ &  41.47$_{\pm 2.32}$ &   66.50$_{\pm 3.02}$ &  70.53$_{\pm 4.13}$ &  84.30$_{\pm 1.48}$ &  62.97$_{\pm 1.40}$ &   78.99$_{\pm 2.43}$ \\
Qwen2.5-7B-Instruct + CoT   &  86.64$_{\pm 2.30}$ &  82.66$_{\pm 3.34}$ &  45.69$_{\pm 0.54}$ &   66.61$_{\pm 1.44}$ &  \underline{78.77}$_{\pm 0.89}$ &  83.41$_{\pm 1.72}$ &  65.94$_{\pm 0.64}$ &   94.56$_{\pm 2.35}$ \\
Gemma-2-9b-it + CoT         &   \textbf{95.72}$_{\pm 0.26}$ &  \underline{83.35}$_{\pm 0.07}$ &  44.89$_{\pm 0.58}$ &   63.46$_{\pm 2.15}$ &  66.07$_{\pm 0.68}$ &  78.40$_{\pm 4.62}$ &  \underline{67.56}$_{\pm 0.64}$ &   67.28$_{\pm 1.04}$ \\
GLM4-9B-Chat + CoT          &   86.58$_{\pm 1.24}$ &  67.01$_{\pm 3.56}$ &  44.49$_{\pm 1.51}$ &   59.72$_{\pm 0.30}$ &  69.34$_{\pm 2.45}$ &  82.06$_{\pm 2.64}$ &  63.21$_{\pm 1.09}$ &   92.91$_{\pm 2.11}$ \\
Mixtral-8x7B-Instruct + CoT &   92.47$_{\pm 0.23}$ &  73.88$_{\pm 2.59}$ &  42.15$_{\pm 4.10}$ &   66.09$_{\pm 2.43}$ &  70.64$_{\pm 0.91}$ &  \underline{84.40}$_{\pm 2.04}$ &  66.71$_{\pm 1.14}$ &   92.01$_{\pm 0.91}$ \\
InternLM-2.5-7B-Chat + CoT  &   89.50$_{\pm 5.76}$ &  69.25$_{\pm 0.95}$ &  39.34$_{\pm 0.09}$ &   49.72$_{\pm 1.52}$ &  64.48$_{\pm 1.21}$ &  80.48$_{\pm 5.42}$ &  56.98$_{\pm 3.97}$ &   76.72$_{\pm 1.66}$ \\
GPT-4o-mini + CoT           &   91.65$_{\pm 2.51}$ &  71.33$_{\pm 0.69}$ &  \underline{46.15}$_{\pm 0.85}$ &   \textbf{68.56}$_{\pm 3.22}$ &  76.27$_{\pm 2.51}$ &  83.63$_{\pm 2.71}$ &  67.16$_{\pm 1.52}$ &   89.22$_{\pm 0.55}$ \\
GPT-4o + CoT                &   \underline{93.80}$_{\pm 3.83}$ &  \textbf{85.50}$_{\pm 2.74}$ &  \textbf{46.92}$_{\pm 1.17}$ &   \underline{67.56}$_{\pm 1.64}$ &  \textbf{84.90}$_{\pm 1.89}$ &  \textbf{87.17}$_{\pm 0.52}$ &  \textbf{76.90}$_{\pm 0.23}$ &   \textbf{95.53}$_{\pm 0.49}$ \\
\bottomrule
\end{tabular}
}
\caption{Results of LLMs on \ours. Bold font and underlining indicate
the best and second-best performance, respectively.}
\label{tab:model_comparison}
\vspace{-6pt}
\end{table*}
\vspace{-5pt}
\section{Experimental setups}
\subsection{Baseline Models}
We evaluate \ours on eight frontier LLMs: Llama-3.1-8B-Instruct~\cite{dubey2024llama}, Qwen-2.5-7B-Chat~\cite{yang2024qwen2}, Gemma-2-9B-it~\cite{team2024gemma}, GLM4-9B-Chat~\cite{glm2024chatglm}, Mixtral-8x7b-Instruct~\cite{jiang2024mixtral}, and ChatGPT-series(GPT-4o-mini, GPT-4o-0806). By following the common practices \cite{kim2023fantom, sabour2024emobench}, we test these models with two types of prompts: (1) vanilla zero-shot prompting directly asks LLMs to give a choice without any explanation; (2) CoT prompting method by following~\cite{kojima2022large} and using the prompt “Let's think step by step.” to elicit the reasoning process and extract the choices by string matching.
The temperature for generating answers is set to 0.7\footnote{LLMs occasionally output with illegal format. We choose a low but nonzero temperature to resample the answers for these invalid generations.}, and we report results across three repetitions.
To measure the specific performance gap between humans and the state-of-the-art machine on the \ours, we employ three graduate students in computer science to complete the human evaluation task. To avoid the bias of LLMs toward a specific choice letter, we shuffle the choices to maintain a nearly uniform distribution of correct choices over the dataset. Prompts used for vanilla zero-shot prompting and CoT prompting are shown in Appendix~\ref{sec: prompts}.
\section{Results and Analysis}
\label{sec: Results}
\subsection{Main Results} The overall evaluation results on \ours for the 8 models are summarized in Table~\ref{tab:model_comparison}, including all the different questions for the persuader and persuadee. We analyze the model's performance for each type of question below.  

\begin{figure*} 
    \vspace{-10pt}
    \begin{minipage}{0.5\columnwidth} 
        \centering
        \includegraphics[scale=0.36]{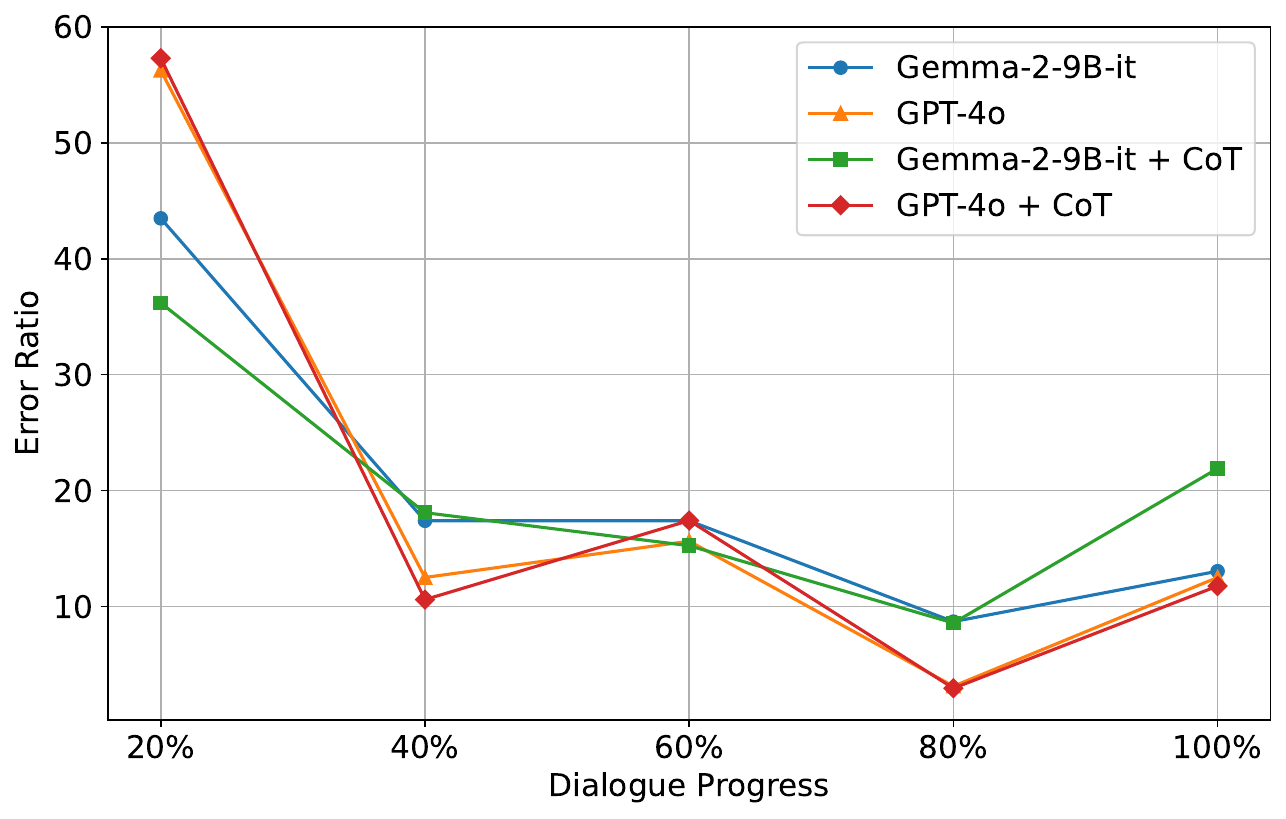} 
        \label{fig:success rate change}
    \end{minipage}
    \hspace{40mm} 
    \begin{minipage}{0.5\columnwidth} 
        \centering
        \includegraphics[scale=0.36]{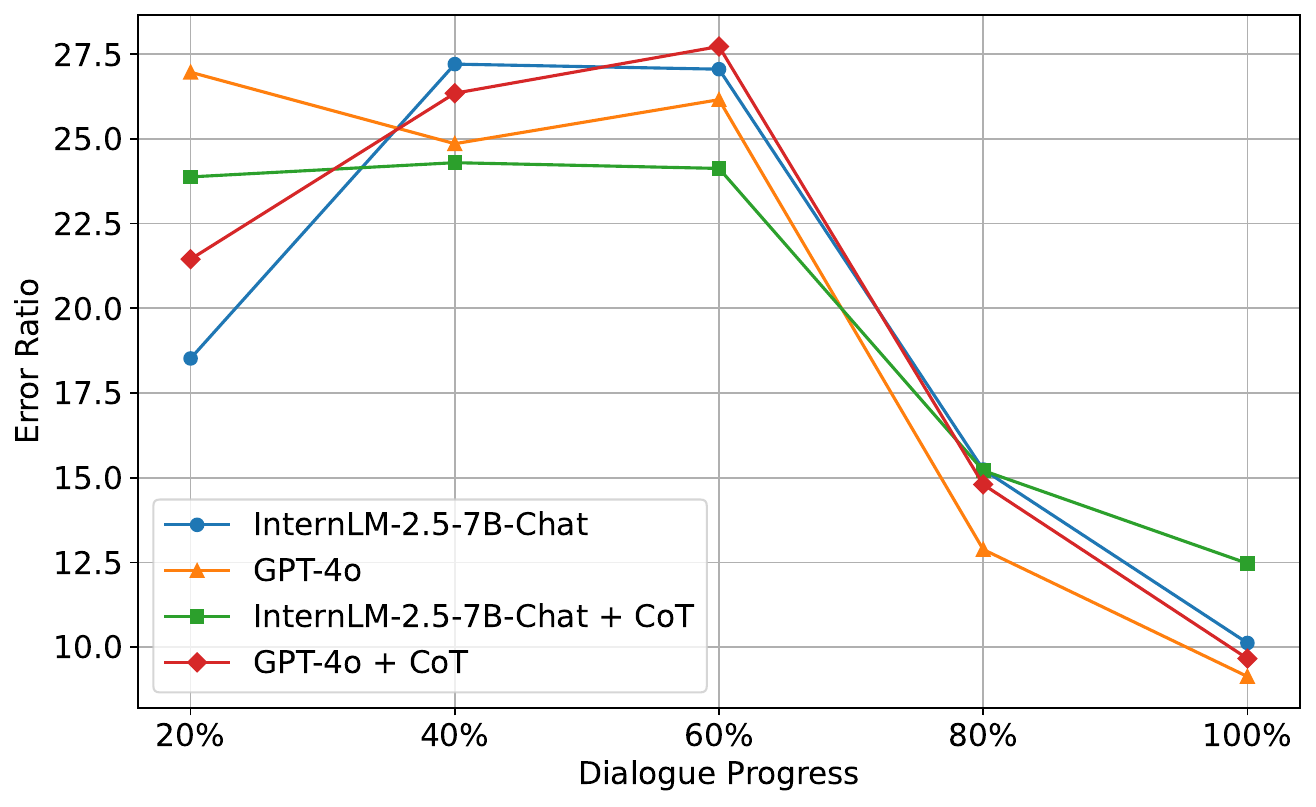} 
        \label{fig:data eff}
    \end{minipage}
    \vspace{-15pt}
    \caption{Distribution of errors of Desire questions happening in different stages of dialogue progress. The \textbf{Left} figure corresponds to the persuader, and the \textbf{Right} figure corresponds to the persuadee.}
    \vspace{-15pt}
    \label{fig: Desire analysis}
\end{figure*}

\paragraph{Desire.} 
Our results show that smaller models, such as Gemma-2-9B and Qwen-2.5-7B, can perform reasonably well in inferring the persuader's desires, achieving an accuracy of over 96\%, which is competitive with GPT-4o. This suggests that most LLMs can easily discern the desires of the persuader. However, when it comes to the desires of the persuadee, performance is relatively lower. Unlike the static desires of the persuader, the persuadee's desires are dynamic, evolving from an initial state to a final state, often with neutral expressions in between. This lower performance highlights that inferring the dynamic desires of the persuadee remains a significant challenge for LLMs. 
\paragraph{Belief.} On belief questions, larger models like GPT-4o perform much better than smaller models on reasoning about the beliefs of both parties. The performance difference between the reasoning persuader's beliefs and the persuadee's beliefs is subtle.
This is because both parties' beliefs dynamically change with each other's speech during the conversation. The difficulty of reasoning persuader and the persuadee's beliefs is similar.
\vspace{-8pt}
\paragraph{Intention.} Results in Table~\ref{tab:model_comparison} indicate that LLMs struggle to accurately infer the intentions of persuaders while performing relatively better at reasoning the intentions of persuadees. The low performance on persuader-related intention questions suggests that LLMs face challenges in understanding how persuaders aim to influence others. This also indicates a lack of proficiency in persuasive theory, limiting the models' ability to correctly interpret and predict the intentions behind the persuader's strategies.

\begin{figure}[t]
\centering
\includegraphics[width=0.88\columnwidth]{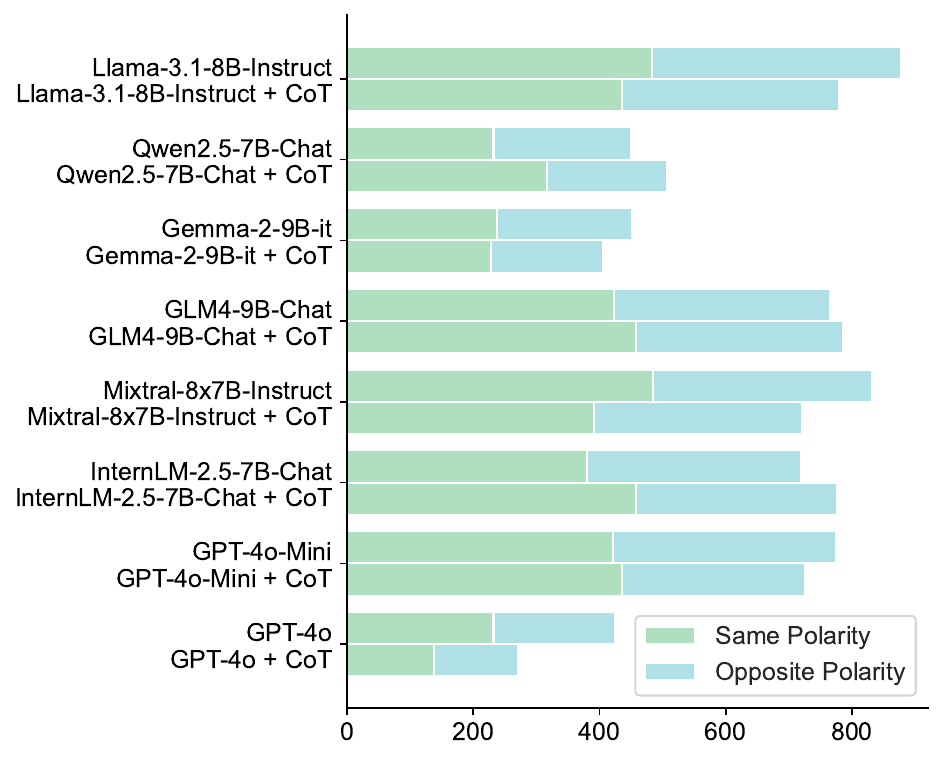}
\vspace{-10pt}
\caption{Model errors of belief questions of persuader.}
\label{fig: belief_er_error}
\vspace{-15pt}
\end{figure}
\vspace{-5pt}
\paragraph{Strategy Prediction and Judgment.} Our results reveal that LLMs perform well in evaluating the effectiveness of persuasive strategies aimed for changing the mental states of the persuadee. However, the task of selecting the appropriate strategy to persuade is more challenging, particularly for smaller models. This suggests LLMs struggle with the complex reasoning required to determine which strategy to adopt in different persuasive contexts.
\vspace{-8pt}
\paragraph{Impact of CoT Reasoning.} Both ToM reasoning and ToM application tasks indicate that CoT reasoning has not consistently improved performance, as observed in ~\cite{kim2023fantom, chen2024tombench}, while it improves strategy prediction to some extent for most LLMs. CoT reasoning involves breaking down the mental states associated with each utterance and generating the mental states of previous utterances. This process can introduce intermediate mistakes that may mislead the overall reasoning about mental states. 
\vspace{-8pt}
\paragraph{Comparison with Human Performance.} To obtain a baseline for human performance, we recruited participants to complete the questions. More details of human evaluation are shown in Appendix~\ref{sec: human eval}. As shown in Table~\ref{tab:model_comparison}, our human participants outperformed LLMs on all tasks. In particular, although GPT-4o reaches close performance in humans, it still falls short of understanding and reasoning the complex dynamics such as the intention of persuaders and the desire of persuadees, which involves complex psychological changes, highlighting a significant gap in current LLMs and humans.

\begin{figure}[t]
\centering
\includegraphics[width=0.88\columnwidth]{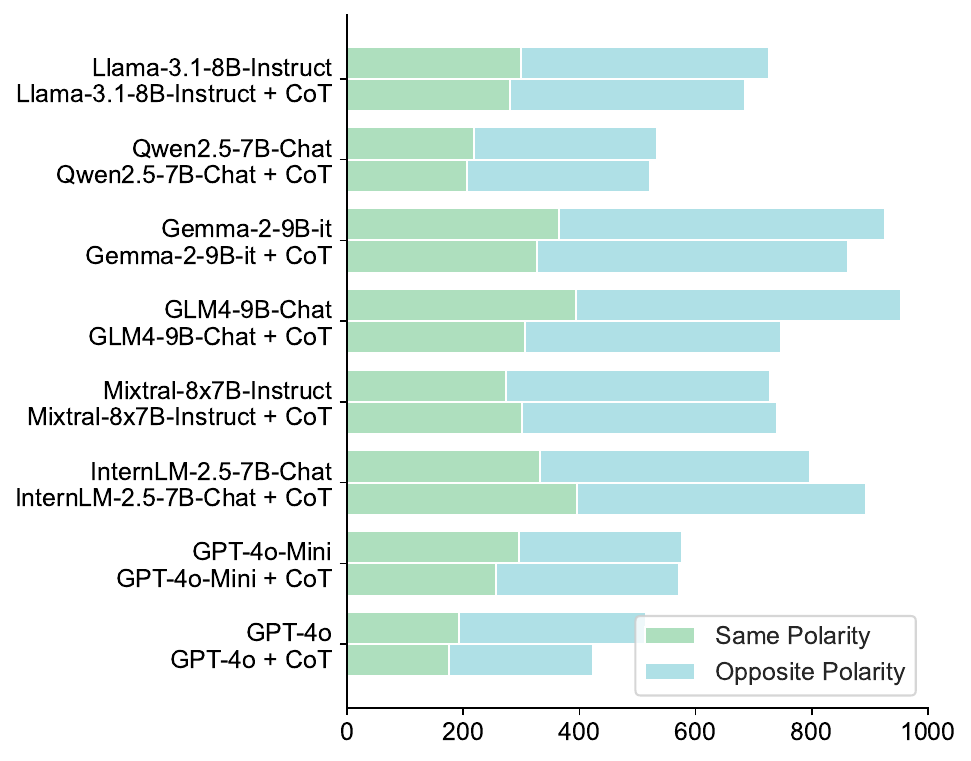}
\vspace{-10pt}
\caption{Model errors of belief questions of persuadee.}
\label{fig: belief_ee_error}
\vspace{-10pt}
\end{figure}
 \vspace{-8pt}
\subsection{In-depth Analysis}
To better understand the limitations of large language models (LLMs) in the \ours benchmark, we categorized common failure cases into several key error types based on task performance and manual error analysis.
\vspace{-6pt}
\paragraph{Desire Reasoning Errors.}
Figure~\ref{fig: Desire analysis} summarizes the distribution of errors of desire questions happening in different stages of dialogue progress with and without CoT reasoning. The error distribution for persuader and persuadee is significantly different. At the beginning of the dialogue, LLMs may not accurately understand the persuader's desire, but as the dialogue progresses, the persuader's desire becomes relatively easy to identify. However, for the persuadee, the desire to reject at the early stage of the dialogue is relatively easy to recognize. As the persuasion proceeds, the persuadee may begin to contemplate and hesitate over the persuader's proposal, leading to complex and nuanced psychological activities that make it difficult for the LLM to accurately judge the persuadee's desire. As the dialogue approaches its end, the persuadee shows a tendency to agree, making the reasoning of desire easier. This suggests LLMs still fall short in ToM reasoning regarding desire shifts.

\begin{figure}[t]
\centering
\includegraphics[width=1.0\columnwidth]{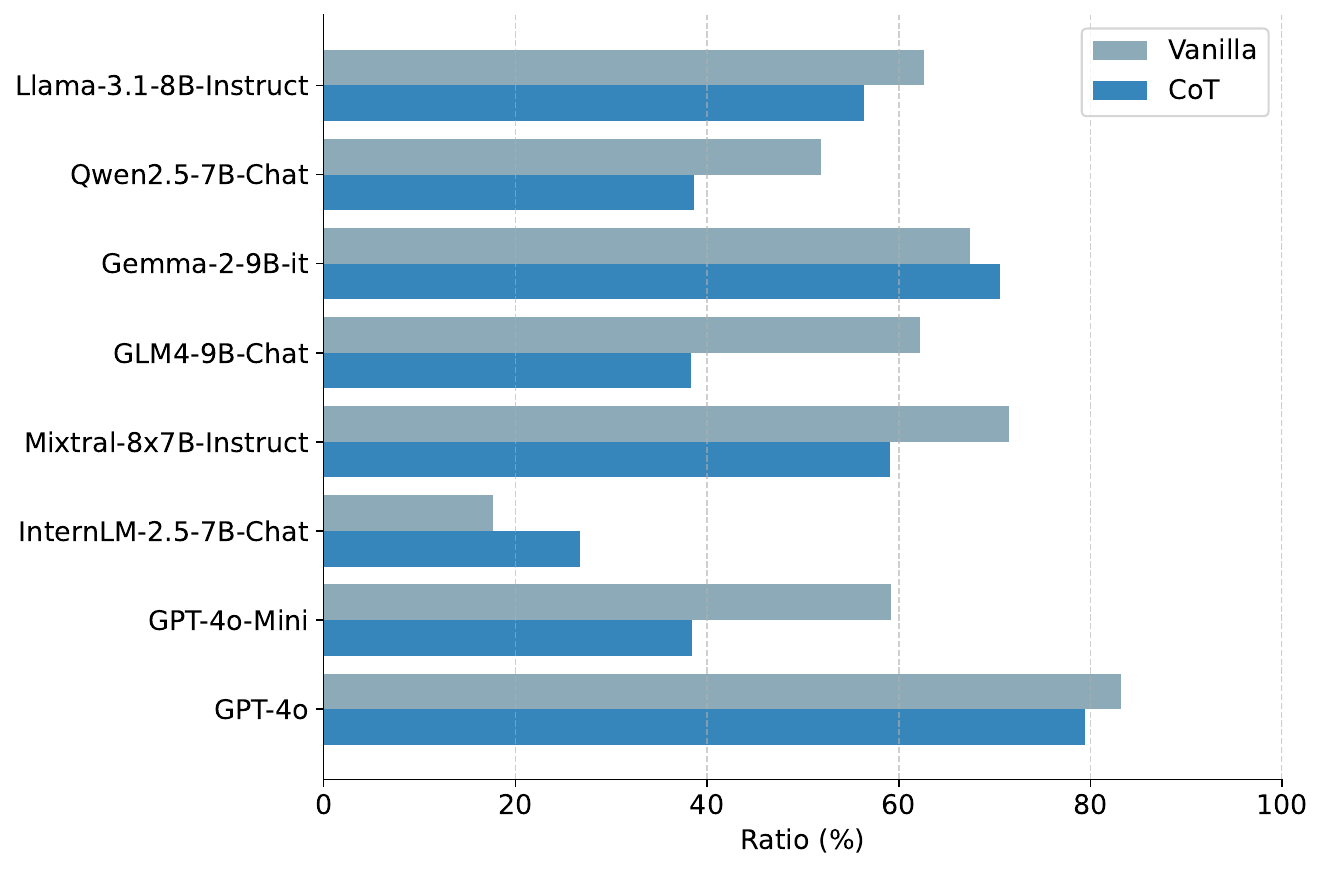}
\caption{Ratio of intention errors misclassified to \textit{feel accepted through concessions, promises, or benefits.}}
\label{fig: intent_er_error}
\vspace{-10pt}
\end{figure}

\paragraph{Belief Reasoning Errors.}
Figure~\ref{fig: belief_er_error} and \ref{fig: belief_ee_error} summarize error types of belief questions for each model with and without CoT. We use Distil-BERT\footnote{https://huggingface.co/distilbert/distilbert-base-uncased-finetuned-sst-2-english}~\cite{sanh2019distilbert} to discriminate whether the choice of LLMs has the same attitude polarity as the ground-truth. we found that LLMs make nearly balanced proportions of same- and opposite-polarity errors in both persuader and persuadee belief reasoning. This balance indicates that LLMs do not adopt a clear or consistent strategy when reasoning beliefs, often guessing without a coherent reasoning framework.
\vspace{-8pt}
\paragraph{Intention Bias.} Given the high error rate in intention questions related to the persuader, we conducted an analysis of the error types. Our findings reveal that most LLMs exhibit a bias toward predicting intentions characterized by \textit{making the other person feel accepted through concessions, promises, or benefits}. Figure~\ref{fig: intent_er_error} illustrates the proportion of errors resulting from misclassifying intentions into this category. We hypothesize that this bias may stem from the pretraining phase, particularly with Reinforcement Learning from Human Feedback (RLHF)~\cite{christiano2017deep}, which tends to prioritize safety and politeness. This may explain the models' bias toward predicting intentions emphasizing benefits and concessions, even when misaligned with the dialogue context. We provide a case study in Appendix~\ref{sec: case}.
\vspace{-8pt}
\paragraph{Discussion.} LLMs struggle with BDI reasoning because these mental states are highly dynamic and context-dependent, making their accurate prediction across a dialogue challenging. To mitigate this issue, future work can explore incorporating external memory mechanisms to track changes in mental states over time. or focus on enhancing the LLM's inherent understanding and reasoning capabilities regarding dynamic psychological states by constructing and training on larger, more richly annotated datasets that capture the nuanced evolution of beliefs, desires, and intentions in various interactive scenarios.
\vspace{-5pt}
\subsection{How Well LLMs Track the Mental States of Persuadees?}
We evaluate whether LLMs can holistically track the persuadee’s mental states throughout the entire dialogue. To assess this, we measure whether the model maintains a coherent understanding across multiple turns, counting a dialogue as successful only if all related questions are answered correctly.

As shown in Table~\ref{tab:consistency}, only a small portion of the dialogues are fully understood, especially those involving desire reasoning. This indicates that LLMs still struggle to track dynamic mental states in persuasive settings, revealing a substantial gap from human performance.

\begin{table}[t]
\centering
\resizebox{0.95\columnwidth}{!}{%
    \begin{tabular}{lccc}
        \toprule
        \textbf{Model} & \textbf{Desire} & \textbf{Belief} & \textbf{Intention} \\
        \midrule
        LLaMa-3.1-8B-Instruct  & 22.31 & 21.71 & 60.76 \\
        LLaMa-3.1-8B-Instruct + CoT & 19.92 & 22.86 & 57.52\\\midrule
        Qwen2.5-7B-Instruct & 19.12 & 31.81 & 56.95 \\
        Qwen2.5-7B-Instruct + CoT & 20.52 & 24.76 & 54.67 \\\midrule
        InternLM2.5 & \textbf{24.70} & 20.19 & 58.10 \\
        InternLM2.5 + CoT & 6.57 & 32.14 & 53.71 \\\midrule
        GPT-4o-mini & \underline{23.39} & 30.77 & 60.79\\
        GPT-4o-mini + CoT & 16.13 & 32.14 & 56.95\\\midrule
        GPT-4o & 19.35 & \underline{36.19} & \textbf{65.71} \\
        GPT-4o + CoT & 6.57 & \textbf{45.38} & \underline{62.02} \\\midrule
        Human & 62.00 & 56.00 & 82.00\\
        \bottomrule
    \end{tabular}
    }
    \caption{The consistency (\%) of the models for ToM reasoning questions of persuadee.}
    \label{tab:consistency}
\vspace{-15pt}
\end{table}
\vspace{-5pt}
\section{Conclusion}
\vspace{-5pt}
This work proposes \ours, a benchmark for evaluating LLMs' ability to reason about complex psychological states and apply them in social decision-making. We conducted extensive experiments and analysis to evaluate the performance of LLMs on the \ours benchmark.

\section*{Limitations}
While \ours offers a comprehensive evaluation of the Theory of Mind in real-life social interaction scenarios within persuasive dialogues, both \ours and previous benchmarks still focus on understanding a character's mental state from the perspective of an observer. However, the ability to reason about others' mental states in persuasive dialogues can further position LLMs as autonomous agents. This capability would enable them to better guide other agents in fulfilling their own desires by reasoning about the mental states of others. Therefore, future benchmarks should establish environments with multiple LLM agents, where tasks involve reasoning about the mental states of other agents and proposing persuasion strategies to influence their desires, beliefs, and intentions to fulfill the current agent’s target. In this context, agents will develop the management skills necessary for effective cooperation and other applications.

\section*{Societal and Ethical Considerations}
We recognize that the concept of the Theory of Mind might suggest anthropomorphic qualities when applied to AI models. However, we want to clarify that our work is not intended to anthropomorphize LLMs. Our goal is to examine the limitations in the social and psychological reasoning capabilities of existing LLMs. Our results show that current models do not perform genuine Theory of Mind reasoning; instead, they generate responses primarily based on the literal interpretation of the input.
\bibliography{custom}
\newpage
\appendix
\section{Additional Discussion}
\label{discussion: dialogue}
\paragraph{Dialogue Geneartion.} \ours is built on multi-turn persuasive dialogues that require models to track changing beliefs, desires, and intentions of both the persuader and persuadee. In practice, LLMs often “get lost” in long conversations~\cite{yang2025large, laban2025llms}. By evaluating how well LLMs can track evolving desires, beliefs, and intentions, \ours provides insights into their capacity for generating contextually appropriate and coherence responses across multiple turns, as demonstrated in Section~\ref{sec: Results}.

\paragraph{Personalized Dialogue.} \ours assesses LLMs' ability to reason about individual mental states, which is essential for tailoring responses to meet specific user needs and preferences~\cite{wang2024unims, chen2024recent}. This understanding allows for more empathetic and effective interactions, enhancing user experience in applications such as customer service and mental health support~\cite{lee2022improving, ma2023towards}. Future work can develop external memory to store user's mental states and retrieve form the memory to generate personalized responses~\cite{huang2024learning}, or internalize them into the LLMs~\cite{li2023learning}.

\section{Data Annotation}
\label{sec: data anno}
\paragraph{Annotation.} In this section, we outline the annotation process and the templates utilized for annotation. Among the various tasks, the intention questions of persuaders and the desire questions of persuadees require annotation. Initially, we recruited three graduate students to annotate 25 dialogues. Subsequently, we carefully designed a few-shot prompt to guide DeepSeek-V3~\cite{liu2024deepseek} as an annotator, aiming to enhance the alignment between the model's answers and human annotations. Following this, we employed the LLM to annotate the remaining questions. To ensure the quality of the annotations, we randomly sampled 100 dialogues and calculated the inter-annotator agreement. The Fleiss $\kappa$ \cite{fleiss1971measuring} was found to be 76.20\% for the desire questions of persuadees and 78.28\% for the intention questions of persuaders. These results are presented in Table~\ref{tab:kappa}, which indicate a high inter-annotator agreement. The detailed statistics and comparison with other ToM datasets are shown in Table~\ref{tab:dataset_stats}.
\paragraph{Choices Generation.}Binary choice questions, including those related to the desires of the persuader and judgment questions, do not require additional choice generation. For belief-related questions for both parties, we adapt the tones from the DailyPersuasion dataset. We also create lists of attitudes—positive, neutral, and negative—and manually remove any items that have semantics too close to the ground truths. From each attitude list, we then randomly sample one word to generate the four choices.

For intention questions concerning the persuader, we directly use the intention options outlined in Table~\ref{intention}. For the persuadee’s intention questions, we leverage DeepSeek-V3 and employ a few-shot prompt (as shown in Figure \ref{fig: extraction}) to extract the persuadee's intention. Subsequently, we design another prompt (as shown in Figure \ref{fig:choice generation}) to generate three incorrect intention choices.

Since DailyDialogue provides persuasion strategies for each utterance, we construct the choices by including the correct strategy and three alternative strategies that appear in other turns of the same dialogue.

\section{Appendix for Experiments}
\subsection{Human Performance}
\label{sec: human eval}
To measure the performance gap between humans and the state-of-the-art LLMs on \ours, we recruited three graduate student workers majoring in computer science to complete the questions. Each question is shown to the workers with identical prompts which are used for evaluating LLMs. We then compute the majority vote on the labels assigned. Student workers solve 50 dialogues in total. For a question where three people have different answers, we randomly select one of their answers as the answer for human evaluation.
\label{sec:appendix}

\subsection{Prompts used for evaluation}
\label{sec: prompts}
Here we show the prompts used for vanilla zero-shot prompting and CoT prompting for generating answers for all the ToM Reasoning and ToM Application questions. We only need to fill in the content to "<>" for evaluating different questions. The vanilla zero-shot prompt is shown in Figure~\ref{fig: vanilla}, and the CoT prompt is shown in Figure~\ref{fig:cot prompting cot}

\begin{table}[t]
    \centering
    \resizebox{0.85\columnwidth}{!}{%
    \begin{tabular}{lc} 
        \toprule
        \textbf{Questions} & \textbf{Fleiss’s Kappa (\%)} \\
        \midrule
        Desire (Persuader)   & 76.20 \\
        Intention (Persuadee) & 78.28 \\
        \bottomrule
    \end{tabular}
    }
    \caption{Inter-rater agreement in terms of Fleiss’s $\kappa$ on desire and intention questions.}
    \label{tab:kappa}
\end{table}

\begin{table*}[t]
\centering
\resizebox{1.0\textwidth}{!}{%
\begin{tabular}{ccccc}
\toprule
\textbf{Dataset} & \textbf{Total \#Questions} & \textbf{Avg. \#Questions per Context} & \textbf{Avg. \#Turns (Full)} & \textbf{Avg. Turn Length} \\\midrule 
ToMi & 6K & 6.0 & 4.9 & 4.7 \\ 
FANToM & 10K & 12.9 & 24.5 & 21.9 \\
NegotiationToM & 13K & 7.0 & 6.0 & 42.2 \\
\ours & 19K & 8.0 & 4.9 & 61.3\\\bottomrule
\end{tabular}
}
\caption{Statistics of \ours and other recent benchmarks.}
\label{tab:dataset_stats}
\end{table*}

\subsection{Case study on Persuader's intention}
\label{sec: case}
Here we present an example of common mistakes made by GPT-4o that misclassifying intentions to \textit{make the other person feel accepted through concessions, promises, or benefits.}, as shown in Table~\ref{table:case_study}. We believe these errors can be attributed to the RLHF which highlights the benefits for humans, as well as the potential unfamilirity of persuasion theory for LLMs.

\section{Details on Persuasive Principles}
\label{sec: persuasion}
Robert Cialdini’s six principles of persuasion, outlined in his book Influence: The Psychology of Persuasion, are foundational concepts in social psychology. They explain how people can be persuaded or influenced by others. We include an overview for each of the principle in Table~\ref{tab: persuasive principle}.

\definecolor{lightGreen}{HTML}{d6fae5}
\definecolor{lightBlue}{HTML}{D9E9F7}
\definecolor{mygreen}{HTML}{14ad56}
\definecolor{myred}{HTML}{cc2424}
\definecolor{myblue}{HTML}{3079BA}
\begin{table*}[t]
  \centering \resizebox{1.0\textwidth}{!}{
    \begin{tabular}{l | l}
        \toprule
        \textbf{Utterance} & \textbf{Bob}: I understand your love for Paris, but Bali also offers a thrilling adventure! We can go \\
        ~ & white water rafting, hike to volcanoes, and explore hidden waterfalls. It's a perfect destination \\ ~ & for creating \textbf{\color{myblue}unforgettable memories} together. \\ 
         \textbf{Question} & What is the intention of Bob?\hspace{0.1cm} \\
         \textbf{GPT-4o} & \textbf{\color{myred}(A) Intent to make the other person feel accepted through concessions, promises, or benefits.} \\
         \textbf{Label} & \textbf{\color{mygreen}(B) Intent to make the other person feel the experience or objects are unique or scarce.}\\
        \bottomrule
    \end{tabular}
    }
  \caption{Common observed mistakes in our experiments. 
  {\color{mygreen} Green} and {\color{myred} Red} indicate the correct answer and GPT-4o's answer, respectively.
  }
  \label{table:case_study}
\end{table*}

\begin{table*}[t]
    \centering
    \resizebox{1.0\textwidth}{!}{%
    \begin{tabular}{l|l}
        \toprule
        \textbf{Principle} & \textbf{Description} \\
        \hline
        Reciprocity principle & Assist others or provide them with gifts, creating a sense of obligation to return the favor.\\
        ~ & For instance, giving away free trials, discount coupons, or complimentary gifts can enhance \\
        ~ & persuasion. \\\midrule
        Scarcity principle & When a resource or opportunity is scarce, people are more inclined to take action. \\
        ~ &  Highlighting urgency and scarcity can motivate the audience to respond quickly.\\\midrule
        Consensus principle & People often follow the actions of others, especially in uncertain situations. Provide \\
        ~ & information such as successful cases of others, positive reviews, or the number of \\
        ~ & supporters to increase persuasiveness.\\\midrule
        Authority principle & People are more likely to trust and follow guidance from authoritative figures. Citing expert \\
        ~ & opinions, research findings, or endorsements from reputable institutions can enhance \\
        ~ & credibility and persuasiveness.\\\midrule
        Commitment and & People are inclined to stick to their past commitments and behave Consistently. \\ 
        consistency principle & Encouraging them to express support or make a small commitment increases the  \\
        ~ & chances of them taking further action later. \\\midrule
        Liking principle & People are more easily influenced by those they like, admire, or find relatable.\\
        ~ & For instance, a salesperson who shares common interests with a customer is \\
        ~ & more likely to make a sale.\\
        \bottomrule
    \end{tabular}
    }
    \caption{Explanations for Robert Cialdini’s six principles of persuasion.}
    \label{tab: persuasive principle}
\end{table*}

\input{fig_texts/desire_ee_anno}

\input{fig_texts/choice_generation}

\input{fig_texts/evaluation}

\input{fig_texts/evaluation_cot}

\end{document}

%% file: fig_texts/desire_ee_anno.tex
\begin{figure*}[t]
  \centering %

\begin{mybox}[Prompt for extracting intention of persuadee.]
\begin{obeylines}
You are a skilled intent understanding expert. You will be given a sentence describing <persuadee's name>'s intent. Please only return the intent without any explanation.
Case 0: 
Sentence: Mary wants excitement, so I'll appeal to her sense of adventure and describe how exploring the ruins can be thrilling. 
Intent: Mary wants excitement 
Case 1: 
Sentence: Oliver is concerned about failure, so discussing the financial benefits of starting an e-commerce business could help alleviate his worries. 
Intent: Oliver is concerned about failure 
Case 2: 
Sentence: Olivia seems intrigued by the idea of personalization. I'll explain how we can incorporate it into our subscription model. 
Intent: Olivia seems intrigued by the idea of personalization. 
Case 3: 
Sentence: <Utterance of persuadee>
Intent: 
\end{obeylines}
\end{mybox}
\vspace{-12pt}
\caption{Prompt template for extracting intention of persuadee.}
\label{fig: extraction}
\end{figure*}

%% file: fig_texts/choice_generation.tex
\begin{figure*}[t]
  \centering %
\begin{mybox}[Prompt for choice generation of intention questions of persuadee]
\begin{obeylines}
You are an expert in multiple-choice question-making. You will be given a correct choice. Please generate three plausible but incorrect choices without any explanation. Only return the incorrect choices for the last case.

Case 0: 
Correct Intent: Mr. Chen needs further persuasion
Analysis: To further persuade Mr. Chen, Li Na should share success stories of other books that have benefited from incorporating literary criticism in their marketing strategies. Providing concrete examples will make her argument more convincing.
Incorrect Intent 1: Mr. Chen is interested in literary criticism.
Incorrect Intent 2: Mr. Chen is looking for success stories.
Incorrect Intent 3: Mr. Chen prefers concrete examples.
Case 1: 
Correct Intent: James is more open to the idea.
Analysis: James is now more open to the idea, so I'll outline the implementation plan and emphasize the program's flexibility to address any concerns about disruptions.
Incorrect Intent 1: James is concerned about disruptions.
Incorrect Intent 2: James is looking for a detailed implementation plan.
Incorrect Intent 3: James is hesitant about the program's flexibility.
Case 2: 
Correct Intent: <correct intent>
Analysis: <analysis>
Incorrect Intent 1: 
Incorrect Intent 2: 
Incorrect Intent 3: 
\end{obeylines}
\end{mybox}
\vspace{-12pt}
\caption{Prompt template for choice generation of intention questions of persuadee.}
\label{fig:choice generation}
\end{figure*}

%% file: fig_texts/evaluation.tex
\begin{figure*}[t]
\centering %

\begin{mybox}[Prompt for vanilla zero-shot prompting.]
\begin{obeylines}
Here is a persuasive dialogue. There are two agents, the persuader and the persuadee. The persuader is trying to persuade the persuadee to do something. Please answer the following questions using A, B, C, D, E, F, without any explanation.

Dialogue History:
<dialogue>

Question:
<Question>
Choices:
<Choice A>
<Choice B>
<Choice C>
<Choice D>
Answer: 
\end{obeylines}
\end{mybox}
\vspace{-12pt}
\caption{Prompt template for vanilla zero-shot prompting.}
\label{fig: vanilla}
\end{figure*}

%% file: fig_texts/evaluation_cot.tex
\begin{figure*}[t]
  \centering %

\begin{mybox}[Prompt for CoT prompting.]
\begin{obeylines}
Here is a persuasive dialogue. There are two agents, the persuader and the persuadee. The persuader is trying to persuade the persuadee to do something. Think step by step to answer the question.

Ending with "The answer is A, B, C, D, E, F". For example, if the most likely answer option is 'A. considering', then end your response with 'The answer is A'. 

Dialogue History:
<dialogue>

Question:
<Question>
Choices:
<Choice A>
<Choice B>
<Choice C>
<Choice D>
Answer: Let's think step by step.
\end{obeylines}
\end{mybox}
\vspace{-12pt}
\caption{Prompt template for CoT prompting.}
\label{fig:cot prompting cot}
\end{figure*}